\documentclass{article}

\usepackage[preprint]{neurips_2025}

\usepackage[utf8]{inputenc} 
\usepackage[T1]{fontenc}    
\usepackage{hyperref}       
\usepackage{url}            
\usepackage{booktabs}       
\usepackage{amsfonts}       
\usepackage{nicefrac}       
\usepackage{microtype}      
\usepackage[table,dvipsnames]{xcolor}
\usepackage{pifont}
\usepackage{amssymb}
\usepackage{bbding}
\usepackage{bm}
\usepackage{amsmath}

\usepackage{tikz}
\usetikzlibrary{arrows.meta}

\usepackage{graphicx}

\usepackage{algorithm}
\usepackage{algpseudocode}

\usepackage{tikz}
\usetikzlibrary{positioning, shapes.geometric, arrows.meta, shadows, calc, backgrounds}

\definecolor{heat1}{RGB}{220, 240, 255} 
\definecolor{heat2}{RGB}{160, 210, 250} 
\definecolor{heat3}{RGB}{60, 120, 220}  
\definecolor{heat4}{RGB}{10, 50, 140}   

\definecolor{maskgray}{RGB}{200, 200, 200} 

\definecolor{tokengreen}{RGB}{220, 255, 220}

\DeclareFixedFont{\ttb}{T1}{txtt}{bx}{n}{9} 
\DeclareFixedFont{\ttm}{T1}{txtt}{m}{n}{9}  
\usepackage{listings}
\definecolor{deepblue}{rgb}{0,0,0.8}
\definecolor{deepred}{rgb}{0.6,0,0}
\definecolor{deepgreen}{rgb}{0,0.5,0}

\usepackage{pythonhighlight}
\usepackage{enumitem}
\usepackage{wrapfig}
\usepackage{multirow}
\usepackage{graphicx}
\usepackage{tcolorbox}
\usepackage{colortbl}
\definecolor{light-gray}{gray}{0.92}
\newcommand{\xhdr}[1]{\noindent{{\bf #1.}}}

\title{
Token Reduction Should Go Beyond Efficiency in Generative Models -- From Vision, Language to Multimodality
}

\author{
    Zhenglun Kong$^1$\thanks{Equal contribution.},
    Yize Li$^2$\footnotemark[1],
  Fanhu Zeng$^3$, Lei Xin$^{4,6}$, Shvat Messica$^1$, \\
  \textbf{Xue Lin$^2$, Pu Zhao$^2$, Manolis Kellis$^5$, Hao Tang$^6$, Marinka Zitnik$^1$}\\
  $^1$Harvard University, 
  $^2$Northeastern University,
  $^3$CAS, \\
  $^4$Wuhan University, 
  $^5$MIT, 
  $^6$Peking University,\\
  \texttt{\{zhenglun\_kong,marinka\}@hms.harvard.edu}, \texttt{li.yize@northeastern.edu}, \\
}

\begin{document}

\maketitle

\begin{abstract}
In Transformer architectures, tokens\textemdash discrete units derived from raw data\textemdash are formed by segmenting inputs into fixed-length chunks. Each token is then mapped to an embedding, enabling parallel attention computations while preserving the input's essential information. Due to the quadratic computational complexity of transformer self-attention mechanisms, token reduction has primarily been used as an efficiency strategy. This is especially true in single vision and language domains, where it helps balance computational costs, memory usage, and inference latency. Despite these advances, this paper argues that token reduction should transcend its traditional efficiency-oriented role in the era of large generative models. In this paper, we characterize this mechanism as a fundamental principle in generative modeling, critically influencing both model architecture and broader applications. We analyze how token reduction addresses critical challenges in current systems across vision, language, and multimodal, demonstrating its ability to: (i) facilitate deeper multimodal integration and alignment, (ii) mitigate overthinking and hallucinations, (iii) maintain coherence over long inputs, and (iv) enhance training stability, etc. We reframe token reduction as more than an efficiency measure. By doing so, we outline future directions for task-aware token reduction, spanning adaptive algorithms, reasoning-centric compression, reinforcement-guided reasoning control, long video understanding, dense prediction, agentic systems, and broader scientific AI domains. \footnote{We collected a list of token reduction papers at: \href{https://github.com/ZLKong/awesome-token-compression-reduction}{\textcolor{blue}{Awesome-Collection-Token-Reduction}}.}

\end{abstract}

\section{Introduction}
Transformer-based generative models~\cite{brown2020language,devlin2019bert,NEURIPS2021_854d9fca,vaswani2017attention,yang2026survey} have emerged as dominant deep learning architectures across vision, language, and multimodal tasks, due to their ability to process long sequences of tokens, which are the fundamental representational units derived from raw data such as subwords in language or image patches in vision. 
As generative models scale to increasingly complex real-world applications, particularly in long-context and agentic settings, sequence lengths grow rapidly. Persistent histories, tool traces, and multi-step interactions further amplify token usage, as recently exemplified by token-intensive agentic systems such as OpenClaw and Codex. 
This growth leads to the quadratic cost of self-attention, substantial memory overhead, and slow inference, thereby posing a central bottleneck for deploying generative models at scale.
Token reduction addresses this challenge by reducing the number of tokens processed during inference. By pruning or merging tokens, token reduction~\cite{guo2025learning,han2025filter,huang2024prunevid,hyun2025multi,Jiang_2025,kim2025faster,li2025mutual,liu2025catanet,wu2025tokenselect,zhang2025cdpruner} reduces computational cost and accelerates runtime, providing a practical solution for enhancing generative efficiency~\cite{cao2026fastdrivevla,feng2026gridsampler,hu2026illava,kim2025faster,liu2026hiprune,liu2025vidcom2,tang2025ufo,wu2025streamline,xing2025vision,xiong2025prune2drive,yao2025timechatonline,zhang2025training,zhang2025vqtoken,zhang2025falcon,zhu2026hawk,zhuang2025vasparse}.

Token reduction has been widely adopted in computer vision, language processing, and multimodal tasks. In vision transformers, it has primarily been used to reduce computational cost by removing visually redundant tokens~\cite{bergner2024tokencrop,bolya2022token,fang2025attend,10.1609/aaai.v39i16.33894,kong2022spvit,Lei_2025_CVPR,liang2022not,rao2021dynamicvit}. 
In language models, token reduction has commonly been implemented through early-exit mechanisms and token-skipping strategies~\cite{lin2025boosting,wu2024accelerating}, which reduce the number of intermediate tokens processed and thus lower computational overhead. Similarly, multimodal large language models~(MLLMs) apply visual token pruning primarily during the prefill stage~\cite{chen2024image}, where adaptive attention patterns are learned in the early layers to prune tokens in later stages.
Despite progress, token reduction is predominantly viewed as a post-hoc efficiency optimization~\cite{liu2025shifting,shao2025tokens},  primarily by reducing the number of tokens to minimize associated computations and accelerate inference. 
Such an efficiency-only mindset has limitations. Naive pruning methods may discard informative tokens, thereby degrading model understanding and performance~\cite{liang2022not,zhan2024exploring,zhan2024rethinkingtoken}. Furthermore, token reduction is commonly treated as a post hoc optimization, rather than being integrated into the core design and training of the model~\cite{chen2024image}.

In this paper, we argue that viewing token reduction purely from an efficiency perspective is fundamentally limited. Instead, we position token reduction as a core design principle in generative modeling, deeply integrated with both training and inference to prioritize tokens that maximize downstream task performance and semantic integrity.

Modern generative tasks present numerous challenges that highlight the need for thoughtful token selection: (i) Ultra-long contexts in language modeling require selective retention of relevant segments to preserve coherence. (ii)  LLMs frequently exhibit overthinking, repeatedly attending to low-value tokens and producing redundant or contradictory outputs. (iii) Multimodal generation tasks often face issues of visual redundancy, where background tokens overshadow salient visual features critical for accurate understanding. (iv)  Noisy or irrelevant tokens introduced during training slow down convergence and harm model stability. 
By learning to intelligently select, merge, or compress tokens based on their contribution to generation objectives, rather than solely on raw redundancy, models can simultaneously reduce computational load, improve robustness, and enhance interpretability and alignment.
This paper makes the following three key contributions:
\begin{itemize}[leftmargin=*]
\item We categorize existing token reduction methods by their functional objective, identifying a transition from efficiency-centric optimizations to task-aware enhancements in vision, language, and multimodal domains.
\item We identify core challenges faced by modern generative models including insufficient visual representation, semantic misalignment, overthinking in reasoning, and training instability. We then demonstrate how principled token reduction strategies can effectively mitigate these issues. 
\item We outline a roadmap for future research on token reduction, including directions for method design, reinforcement learning-guided token selection, adaptive in-context compression, and hardware-algorithm co-design, etc. These directions aim to support the development of next-generation generative architectures that are both robust and efficient.
\end{itemize}

This position paper is organized as follows: Sec.~\ref{sec:relate} reviews prior token reduction methods across various modalities.  Sec.~\ref{sec:form} introduces the problem formulation, Sec.~\ref{sec:challenge} formalizes the identified challenges and demonstrates how informed token reduction strategies can address them. Sec.~\ref{sec:future} proposes promising research directions for advancing token reduction as well as broader implications.
\section{Related Work}
\label{sec:relate}

\subsection{Token Reduction in Vision Models}
\xhdr{Image Classification} Classification serves as a fundamental task for vision models and token reduction techniques have been widely applied in it due to its simplicity and versatility. It has been widely explored from various aspects~\cite{bolya2022token,liang2022not,rao2021dynamicvit,wu2023ppt,zeng2024m2m,kong2023peeling,zeng2025token}. Specifically,  DynamicViT~\cite{rao2021dynamicvit} devises a lightweight module to predict the importance score of each token, thereby pruning unimportant tokens. SPViT~\cite{kong2022spvit} introduces a soft pruning technique, which integrates the less informative tokens generated by the selector module into a package token that will participate in subsequent calculations rather than being completely discarded. EViT~\cite{liang2022not} identifies attentive tokens from the attention map, enabling token pruning without additional parameters. 
ToMe~\cite{bolya2022token} merges tokens with similarity based on bipartite matching to maintain information utility. PPT~\cite{wu2023ppt} analyzes the statistic data between layers and adaptively employs token pruning and merging within layers to achieve higher acceleration performance.

\definecolor{VisionColor}{HTML}{1F77B4}   
\definecolor{LanguageColor}{HTML}{2CA02C} 
\definecolor{MLLMColor}{HTML}{995F2F}     
\definecolor{VLAColor}{HTML}{FF7F0E}      
\definecolor{AgentColor}{HTML}{D62728}    
\definecolor{TrendGray}{HTML}{555555}

\newcommand{\TRVision}[1]{\textcolor{VisionColor}{#1}}
\newcommand{\TRLanguage}[1]{\textcolor{LanguageColor}{#1}}
\newcommand{\TRMLLM}[1]{\textcolor{MLLMColor}{#1}}
\newcommand{\TRVLA}[1]{\textcolor{VLAColor}{#1}}
\newcommand{\TRAgent}[1]{\textcolor{AgentColor}{#1}}

\newcommand{\LegendItem}[2]{%
  \raisebox{0.6pt}{\textcolor{#1}{\rule{0.55cm}{2.0pt}}}~\textbf{#2}%
}
\begin{figure*}[t]
  \centering
  \begin{tikzpicture}[
      x=0.35cm,
      y=0.6cm,
      axis/.style      = {thick, ->, >={Stealth[length=3pt,width=2pt]}},
      yearline/.style  = {gray!35, thick},
      tick/.style      = {thick},
      marker/.style    = {circle, inner sep=2.8pt},
      event/.style     = {font=\scriptsize, align=left, inner sep=2pt},
      event_top/.style = {event, anchor=south west},
      event_bot/.style = {event, anchor=north west},
      legend/.style    = {font=\footnotesize, align=left, anchor=west, inner sep=1pt},
      trendVL/.style   = {thick, ->, >={Stealth[length=3pt,width=2pt]}, draw=TrendGray!85},
      trendMLLM/.style = {thick, ->, >={Stealth[length=3pt,width=2pt]}, draw=MLLMColor!90!black},
      trendVA/.style   = {thick, ->, >={Stealth[length=3pt,width=2pt]}, draw=VLAColor!90!black}
    ]

    \useasboundingbox (-0.8,-5.85) rectangle (39.8,4.85);

    \draw[axis] (-1.0,0) -- (40.0,0)
      node[pos=0.985, below=4pt, font=\footnotesize]{\textbf{Year}};

    \coordinate (Y2021) at (0.2,0);
    \coordinate (Y2022) at (2.2,0);
    \coordinate (Y2023) at (12.4,0);
    \coordinate (Y2024) at (14.6,0);
    \coordinate (Y2025) at (21.8,0);
    \coordinate (Y2026) at (27.5,0);

    \foreach \year in {2021,2022,2023,2024,2025,2026}{
      \draw[tick] (Y\year) -- ++(0,-0.15);
    }


    \node[event_bot, text width=4cm] (E2021) at ([yshift=-0.15cm]Y2021) {
      \textbf{2020--2021}\\[-1pt]
      \TRLanguage{Power-BERT \cite{goyal2020power}}
      \TRVision{DynamicViT \cite{rao2021dynamicvit}}\\
      \TRVision{TokenLearner \cite{ryoo2021tokenlearner}}
      \TRLanguage{SpAtten \cite{wang2021spatten}}
    };

    \node[event_top, text width=5.95cm] (E2022) at ([yshift=0.15cm]Y2022) {
      \textbf{2022}\\[-1pt]
      \TRVision{LTP \cite{kim2022learned}}
      \TRVision{SPViT \cite{kong2022spvit}}\\
      \TRVision{STTS \cite{wang2022efficient}}
      \TRVision{EViT \cite{liang2022not}}\\
      \TRVision{PatchSlim \cite{tang2022patch}}
      \TRVision{PPT \cite{ma2022ppt}}
    };

    \node[event_bot, text width=5.05cm] (E2023) at ([yshift=-0.15cm]Y2023) {
      \textbf{2023}\\[-1pt]
      \TRVision{ToMe \cite{bolya2022token}}
      \TRMLLM{PuMer \cite{cao2023pumer}}\\
      \TRLanguage{LLMLingua \cite{jiang2023llmlingua}}
      \TRVision{DTP \cite{nawrot2022efficient}}\\
      \TRVision{TPS \cite{wei2023joint}}
      \TRVision{ToMeSD \cite{bolya2023token}}
    };

    \node[event_top, text width=5.95cm] (E2024) at ([yshift=0.15cm]Y2024) {
      \textbf{2024}\\[-1pt]
      \TRMLLM{Honeybee \cite{cha2024honeybee}}
      \TRMLLM{FastV \cite{chen2024image}}\\
      \TRMLLM{LLaMA-VID \cite{li2024llama}}
      \TRMLLM{TokenPacker \cite{li2024tokenpacker}}\\
      \TRMLLM{BRAVE \cite{kar2024brave}}
      \TRVision{DyDiT \cite{zhao2024dynamic}}
      \TRLanguage{TALE \cite{han2024token}}
    };

    \node[event_bot, text width=6cm] (E2025) at ([yshift=-0.15cm]Y2025) {
      \textbf{2025}\\[-1pt]
      \TRMLLM{DivPrune \cite{alvar2025divprune}}
      \TRMLLM{SeTok \cite{wu2024towards}}
      \TRMLLM{Dycoke \cite{tao2025dycoke}}\\
      \TRMLLM{VisionZip \cite{yang2025visionzip}}
      \TRMLLM{SparseVLM \cite{zhang2024sparsevlm}}
      \TRMLLM{TopV \cite{yang2025topv}}\\
      \TRMLLM{VoCo-LLaMA \cite{ye2024voco}}
      \TRVLA{EfficientVLA \cite{yang2025efficientvla}}
      \TRAgent{S$^2$-MAD \cite{zeng2025s}}
    };

    \node[event_top, text width=8cm] (E2026) at ([yshift=0.15cm]Y2026) {
      \textbf{2026}\\[-1pt]
      \TRVLA{FastDriveVLA \cite{cao2026fastdrivevla}}
      \TRVLA{Prune2Drive \cite{xiong2025prune2drive}}\\
      \TRVLA{CoLaVLA \cite{peng2025colavla}}
      \TRVLA{Grid Sampler \cite{feng2026gridsampler}}\\
      \TRAgent{AgentConductor \cite{wang2026agentconductor}}
      \TRAgent{LatentMAS \cite{zou2025latent}}\\
      \TRAgent{SWE-Pruner \cite{wang2026swe}}
    };

    \begin{pgfonlayer}{background}
      \draw[yearline] (Y2021) -- (Y2021 |- E2021.south);
      \draw[yearline] (Y2022) -- (Y2022 |- E2022.north);
      \draw[yearline] (Y2023) -- (Y2023 |- E2023.south);
      \draw[yearline] (Y2024) -- (Y2024 |- E2024.north);
      \draw[yearline] (Y2025) -- (Y2025 |- E2025.south);
      \draw[yearline] (Y2026) -- (Y2026 |- E2026.north);
    \end{pgfonlayer}

    \node[marker, fill=VisionColor!40] at (Y2021) {};
    \node[marker, fill=VisionColor!55] at (Y2022) {};
    \node[marker, fill=MLLMColor!60]   at (Y2023) {};
    \node[marker, fill=MLLMColor!75]   at (Y2024) {};
    \node[marker, fill=VLAColor!65]    at (Y2025) {};
    \node[
      marker,
      fill=VLAColor!78,
      draw=AgentColor!95!black,
      line width=0.7pt
    ] at (Y2026) {};

    \node[legend] at (4.0,-3.15) {
      \LegendItem{VisionColor}{Vision}
      \hspace{0.55em}
      \LegendItem{LanguageColor}{Language}
      \hspace{0.55em}
      \LegendItem{MLLMColor}{Multimodal LLM}
      \hspace{0.55em}
      \LegendItem{VLAColor}{VLA}
      \hspace{0.55em}
      \LegendItem{AgentColor}{Agents}
    };

    \draw[trendVL, line width=1.6pt,>={Stealth[length=4pt,width=7pt]}] (0.0,3.39) --
      node[midway, above=1pt, font=\footnotesize, text=TrendGray!95]
      {Vision \& Language Transformer}
      (17.4,3.39);

    \draw[trendMLLM, line width=1.6pt,>={Stealth[length=4pt,width=7pt]}] (18.1,3.39) --
      node[midway, above=1pt, font=\footnotesize, text=MLLMColor!90!black]
      {LLM \& Multimodal LLM}
      (28.8,3.39);

    \draw[trendVA, line width=1.6pt,>={Stealth[length=4pt,width=7pt]}] (29.4,3.39) --
      node[midway, above=0.1pt, font=\footnotesize, text=VLAColor!90!black]
      {VLA \& Agents}
      (38.7,3.39);

    \node[anchor=west, inner sep=0pt] at ($(Y2026)+(0,5.08)$) {%
      \resizebox{3.75cm}{!}{%
        \textcolor{AgentColor!58!black}{\textbf{Efficiency}}%
        \textcolor{green!60!black}{$\,\bm{\checkmark}\,$}%
        \textcolor{VLAColor!95!black}{$\bm\Rightarrow$}%
        \textcolor{AgentColor!78!black}{\,\textbf{Future}}%
        \textcolor{AgentColor}{$\,\bm{?}$}%
      }%
    };

  \end{tikzpicture}
  \vspace{-15.5mm}

  \caption{Timeline of representative token reduction works with modality- and task-aware color coding, highlighting the recent transition (Vision \& Language Transformers $\to$ LLMs \& MLLMs $\to$ VLA \& Agentic systems).}
  \label{fig:ml_timeline}
\end{figure*}
\xhdr{Video Compression} Unlike token reduction in image classification, video compression focuses more on the temporal redundancy within videos, and algorithms are developed to reduce the number of tokens with less computational overhead. 
Various token reduction methods have been investigated for different tasks, including video understanding~\cite{ryoo2021tokenlearner,sun2025llavascissortokencompressionsemantic,wang2022efficient}, video editing~\cite{li2024vidtome}, video-text retrieval~\cite{liu2022ts2, shen2024tempme}, video action detection~\cite{chen2023efficient}, and so on. 
Specifically, STTS~\cite{wang2022efficient} introduces a lightweight framework that dynamically selects the most informative spatial-temporal tokens in video transformers. Tokenlearner~\cite{ryoo2021tokenlearner} proposes an adaptive tokenization module that learns a handful of informative spatial-temporal tokens, significantly reducing computational costs. EVAD~\cite{chen2023efficient} selectively drops irrelevant spatial-temporal tokens in non-keyframes while preserving keyframe and motion-relevant tokens, and then refines actor features using a context-aware decoder to maintain accuracy with reduced computations.

\xhdr{Generative Tasks} Token reduction in generative tasks~\cite{ju2024turbo} aims to accelerate generative models through 
the efficient utilization of tokens. 
It can be applied to both diffusion models~\cite{bolya2023token,lu2025hdc,wang2024attention} and diffusion transformers~\cite{gao2023masked,zou2025accelerating}.
Specifically, ToMeSD~\cite{bolya2023token} exploits natural redundancy in generated images by merging redundant tokens, successfully extending token merging to stable diffusion with simple unmerging. DyDiT~\cite{zhao2024dynamic} 
reduces redundancy with a Timestep-wise Dynamic Width approach to adopt model width conditioned on the generation timesteps, and a Spatial-wise Dynamic Token  strategy to avoid redundant computations at unnecessary spatial locations.

\subsection{Token Reduction in Language Models}
Token reduction strategies in language modeling have evolved from early optimizations for BERT~\cite{huang2022pyramid,kim2020length,kim2022learned,kim2023leap,ye2021tr} to techniques specifically designed for LLMs. 
PoWER-BERT~\cite{goyal2020power} introduces progressive word-vector elimination by removing redundant token representations based on self-attention dynamics, improving inference efficiency. Learned token pruning~\cite{kim2022learned} extends this approach by learning attention-based thresholds to adaptively prune uninformative tokens, thereby reducing computational costs while preserving model performance.
In LLMs, token reduction must account for the constraints of autoregressive decoding across diverse downstream tasks. Dynamic pooling methods~\cite{anagnostidis2023dynamic,tao2025saliency} adjust token representations on the fly during inference to reduce redundancy. Prompt compression techniques~\cite{fu2024lazyllm,ge2023context,jiang2023llmlingua} aim to reduce computational overhead by compressing the input prompt before generation. Selective decoding approaches~\cite{fu2024lazyllm,wingate2022prompt} reduce per-step inference costs by computing key-value pairs only for tokens critical to predicting the next token.
In multi-agent systems, recent works increasingly view token reduction as a system-level mechanism for managing memory, communication, tool observations, and long-horizon interaction traces~\cite{zeng2025s,kang2025acon,zou2025latent,agentslimming2026,feng2026agentocr}. For example, S$^2$-MAD~\cite{zeng2025s} proposes a sparsification mechanism that limits unnecessary token exchanges between agents, reducing communication costs and improving the efficiency of collaborative reasoning.






\subsection{Token Reduction in Multimodal LLMs}

Recent work has explored visual token pruning by addressing attention inefficiencies of deep transformer layers in MLLMs~\cite{alvar2025divprune,arif2025hired,lin2025boosting,shang2024llava}. Specifically, FastV~\cite{chen2024image} shows that deeper vision-language layers expend significant computations on redundant image tokens. To address this, a lightweight module is adopted to adaptively prune these tokens, reducing inference overheads in subsequent stages.
A complementary approach modifies vision feature extractors or projectors to output a smaller set of highly informative image tokens, effectively distilling the input into a compressed representation~\cite{cai2024matryoshka,cha2024honeybee,kar2024brave,li2024tokenpacker,li2024mini,ye2024voco}. 
However, efficiency gains from these prefill stages 
often fade during the decoding phase, where per-token computations dominate. To overcome this, recent methods jointly optimize token reduction during both prefill and decoding stages, ensuring sustained speedups throughout inference~\cite{huang2024dynamic,song2024less}. Furthermore, the scope of token reduction has recently expanded to Vision-Language-Action (VLA) models, optimizing efficiency for real-time robotic manipulation and autonomous driving~\cite{cao2025fastdrivevla,pei2025action,yang2025efficientvla,jiang2025better}.

In Fig.~\ref{fig:ml_timeline}, we present a timeline of notable developments in token reduction methods, illustrating the shift from early applications in ViT and BERT-based models to more recent advances in LLMs, MLLMs and Agent systems.

\section{Problem Formulation}
\label{sec:form}


In modern generative models~\cite{Qwen2.5VL,2024llama3,liu2023llava,openai2024gpt4,Peebles2022DiT}, a token denotes one fundamental unit of input or representation, typically encoded as a vector. For example, a token might correspond to a subword in language, a patch in an image, or an embedding of a time step in audio. We denote a sequence of $N$ input tokens as $X = [x_1,\dots,x_N] \in \mathbb{R}^d$. 
Token reduction refers to any operation that compresses the token sequence to $M$ tokens~(with $M < N$) by removing or consolidating tokens while aiming to preserve the original information. 

Broadly, token reduction methods fall into four categories: 1) Token pruning methods~\cite{bolya2022token} that remove entire unimportant tokens, simply dropping them from the sequence; 2) Token merging methods~\cite{bolya2022token,bolya2023token} which fuse information from multiple tokens into fewer tokens, effectively compressing the sequence by merging similar or related tokens; 3) Hybrid strategies~\cite{cao2023pumer,kim2024token,wu2023ppt} that combine pruning and merging within a unified framework; 4) Token distillation approaches~\cite{cai2024matryoshka,mu2023learning} which integrate rich information across longer input sequences or multiple modalities into fewer condensed tokens, enabling efficient cross-modal interactions and long-context reasoning in LLMs and MLLMs. 

A core challenge in token reduction is the determination of  tokens to be pruned or merged. There are various importance criteria and scoring mechanisms to rank token significance, including 
attention-based heuristics~\cite{liang2022not}, gradient or loss-based criteria~\cite{huang2024dynamic}, clustering~\cite{haurum2024agglomerative}, and learned predictors~\cite{rao2021dynamicvit}.

From a purely efficiency-oriented perspective, token reduction delivers substantial computational efficiency gains by reducing the quadratic computation cost from $\mathcal{O}(N^{2})$ to $\mathcal{O}(M^{2})$ in attention mechanisms. 
By eliminating redundant tokens and processing fewer computations during inference,  it effectively accelerates the inference speed and improves the model throughput, which is crucial for latency-sensitive tasks or real-time applications.
Furthermore, it reduces the memory footprint of activations and gradients~(e.g., key/value caches), alleviating memory usage for both inference and training, which is particularly beneficial for wide-scale deployments on resource-limited platforms. 
We present more theoretical detail in the Appendix~\ref{sec:appendix_methodology}

However, as stated in this position paper, token reduction can benefit models in multiple ways beyond efficiency, which will be introduced in detail in the following sections.

\section{Core Roles and Challenges}
\label{sec:challenge}
In this section, we discuss token reduction as a foundational mechanism for addressing critical challenges in modern generative systems. We categorize five core challenges across modalities: visual representation sparsity, semantic misalignment, reasoning redundancy, training instability, and long-context overload. We demonstrate how principled token reduction strategies intrinsically address these issues through dynamic token-semantic co-optimization. We position token reduction not only as an efficiency tool, but as an essential paradigm for enhancing semantic coherence and enabling sustainable scaling of generative systems.
\subsection{Obtain Informative Visual Representation}
MLLMs often suffer from noisy visual inputs that impede fine-grained understanding. We outline key challenges in MLLM visual reasoning:~\ding{172} \textit{Text-Visual Attention Shift:} Due to the rotary positional embeddings in LLM decoders, later text tokens disproportionately attend to spatially lower image regions~\cite{hong2024token}, shifting attention away from semantically important areas~(e.g., objects at the top of an image);~\ding{173} \textit{Visual Redundancy:} Empirical studies~\cite{lin2025boosting,wu2024accelerating} show that beyond the first few layers, many image tokens contribute little new information,~\ding{174} \textit{Task-Guided Focus in VQA:} In multimodal question answering, the question itself pinpoints relevant image regions~(e.g., "kitten color" directs focus to the kitten patch), implying that many image tokens are unnecessary for correct answers~\cite{song2024less}.

\begin{figure*}[t]
\vspace{-0.4cm}
  \centering
  \includegraphics[width=1.0\linewidth]{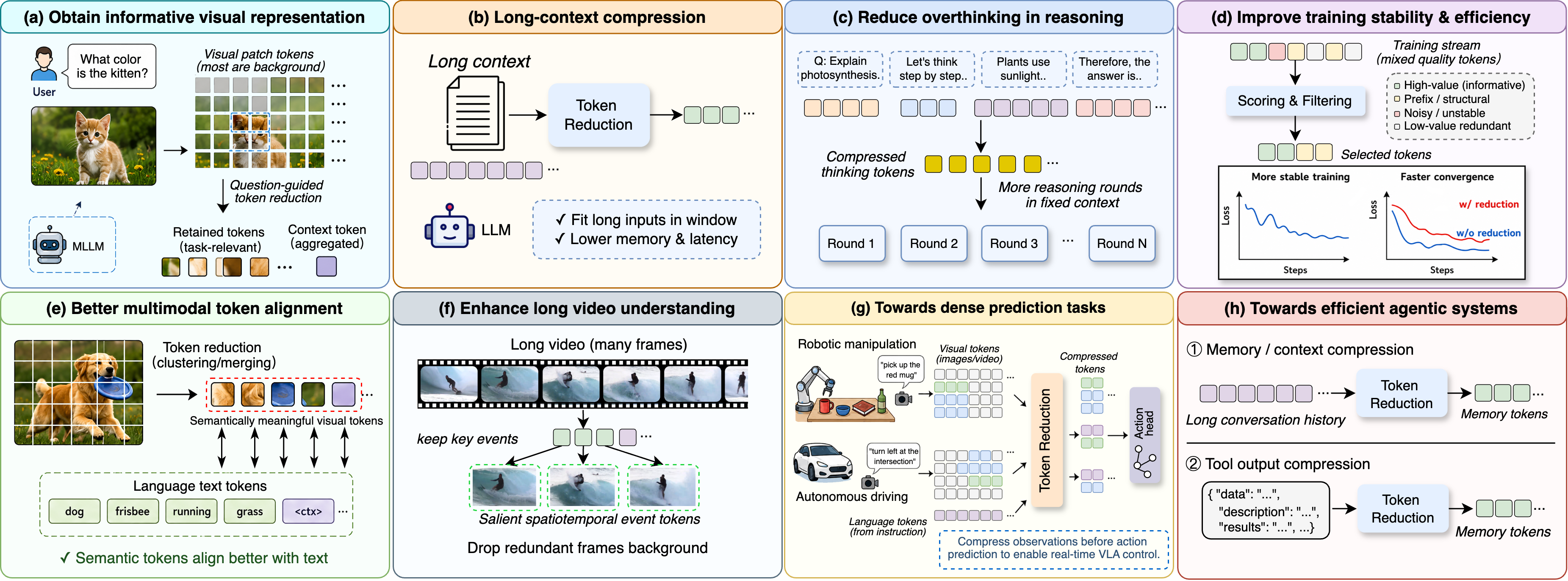}
  \vspace{-0.7cm}
  \caption{Representative roles and application scenarios of token reduction in both vision and language domains. }
  \label{fig:main_framework}
  \vspace{-0.4cm}
\end{figure*}

Therefore, token reduction can serve as a representation-learning optimization: selecting the subset of tokens that preserves informative visual representation, as shown in Fig.~\ref{fig:main_framework}a.  For example, VisPruner~\cite{zhang2025textvisualattentionexploitingvisual} identifies high-value tokens using visual-encoder attention and removes duplicates via clustering to ensure diversity. VTW~\cite{lin2025boosting} observes that visual information migrates into text tokens within early layers; it therefore withdraws all visual tokens after a chosen layer based on KL-divergence criteria.
TRIM~\cite{song2024less} leverages the CLIP metric and IQR scoring function to adaptively select image tokens that are crucial for answering questions, while an aggregated token is used to retain additional image information.

\subsection{Better Multimodal Token Alignment}

Despite impressive capabilities, MLLMs continue to face challenges in semantic alignment. Standard vision tokenizers typically split images into fixed-size patches, which can fragment coherent visual entities~(e.g., objects or regions) across multiple tokens. This fragmentation weakens the alignment between visual and linguistic representations.
Token reduction offers a promising solution by selecting visual tokens based on semantic importance, thereby producing a compact set of tokens that better align with language representations, as shown in Fig.~\ref{fig:main_framework}e. Specifically, SeTok~\cite{wu2024towards} dynamically clusters visual features into semantically meaningful tokens using a density-peak algorithm, which determines the structure of token groupings per image. This approach preserves both high- and low-frequency semantics, substantially improving concept-level alignment and downstream task performance. M3~\cite{cai2024matryoshka} introduces a hierarchical token structure that captures coarse-to-fine semantic granularity, allowing different levels of abstraction to be selectively retained depending on task needs.

\subsection{Reduce Overthinking in Reasoning}
\xhdr{LLM reasoning} In the context of language models, overthinking refers to generating excessively long or convoluted chains of reasoning that go beyond what is necessary to reach a correct answer. An LLM may produce verbose, repetitive, or even self-contradictory explanations when it fails to converge on a solution-often due to uncertainty~\cite{sui2025stop,wang2025harnessing}. Such extended reasoning trajectories are inefficient and recent studies show that state-of-the-art reasoners can consume over 15,000 tokens to solve math problems that could be addressed with a concise chain-of-thought~(CoT) of just a few hundred tokens~\cite{hou2025thinkprune}. This issue is particularly acute in LLM agents, where internal reasoning alternates with external tool use~\cite{gao2025txagent,wang2024toolgen}; excessive steps can obscure logical clarity and lead to error accumulation. 
Mitigating overthinking is thus crucial. By trimming unnecessary tokens, LLMs can focus on salient steps, aligning generation with a more concise trajectory, as shown in Fig.~\ref{fig:main_framework}c.

CoT-Influx~\cite{huang2023fewer} introduces a CoT pruning strategy in which concise reasoning examples are included in the prompt. By pruning unimportant tokens from these examples, more reasoning demonstrations can fit into the context window, surprisingly leading to improved math reasoning accuracy. TokenSkip~\cite{xia2025tokenskip} enables LLMs to skip less important tokens within CoT sequences and learn shortcuts between critical reasoning steps. This allows for controllable CoT compression with adjustable compression ratios, enabling models to automatically trim redundant tokens during reasoning.

\xhdr{MLLM reasoning} MLLMs, which reason over text and other modalities, face 
similar overthinking issues. In vision-language tasks, overthinking often manifests as excessive processing of visual tokens or overly detailed image descriptions, resulting in inefficiency and potential confusion~\cite{liu2026videoauto}. Token reduction techniques in MLLMs aim to promote more focused and sparse reasoning over multimodal inputs. For example, FAST~\cite{xiao2025fast} rewards shorter-than-average token sequences for correct answers, while allowing longer reasoning for more complex tasks. It also adjusts policy optimization constraints to tighten output exploration for simple tasks~(thus reducing unnecessary tokens) and loosen it for harder ones to allow deeper reasoning. 

Together, these strategies reduce overthinking in straightforward cases, boosting efficiency while preserving effective reasoning depth for complex scenarios.



\subsection{Improve Training Stability \& Efficiency}
While token reduction has traditionally been employed as a post-training optimization to enhance inference efficiency, recent research indicates its potential to significantly improve training stability when integrated into the pre-training phase~\cite{gao2023masked,li2024pruning,lin2024not}, suggesting that selective token utilization during training can lead to more robust model learning, as shown in Fig.~\ref{fig:main_framework}d. 

One notable approach is Rho-1~\cite{lin2024not}, which involves scoring tokens based on their alignment with a desired distribution using a reference model and then focusing the training loss on tokens with higher scores. Therefore, it effectively filters out noisy or less informative tokens, leading to faster convergence and improved performance. UPFT~\cite{ji2025first} emphasizes the importance of initial reasoning steps in training. By reducing the number of training tokens,  UPFT encourages the model to focus on the initial prefix substrings of reasoning trajectories, which are often more stable and contain crucial information. This focus helps the model avoid being influenced by subsequent complex or potentially erroneous information, thereby improving training stability.

Additionally, integrating token reduction with training procedures like GRPO~\cite{shao2024deepseekmath} is gaining traction; for instance, recent work reveals that optimizing only a subset of high-entropy ''forking tokens'' matches full-gradient updates~\cite{wang2025beyond}, suggesting that entropy patterns can effectively guide efficient policy learning.
Future research should investigate specialized approaches that incorporate token reduction directly into training objectives, enabling models to learn to prioritize or discard tokens in a task-aware and gradient-aligned manner.

\subsection{Enhance Long Context \& Video Understanding}

\xhdr{Long-context LLMs} Long-context language modeling presents unique challenges:~\ding{172} Long texts often contain raw tokens that exhibit repetitive descriptions and irrelevant details that strain the attention mechanism;~\ding{173} LLM-based agent systems use input data as sequential prompts for reasoning or for switching between multiple tasks, which can lead to overload when the prompt grows too large;~\ding{174} It is very difficult to scale up to even longer content for learning more information. Token reduction techniques directly address these issues by distilling extensive input sequences into compact summary vectors or representative tokens, as shown in Fig.~\ref{fig:main_framework}b. By doing so, models preserve core information such as key events, central themes, or task-specific facts, while decreasing cognitive load. 
For example, 
AutoCompressors~\cite{chevalier2023adapting} trains pre-trained LLMs to compress long contexts into compact summary tokens, reducing token length by orders of magnitude to extend context windows and speed up inference. TokenSwift~\cite{wu2025hours} reduces the effective number of tokens that the model dynamically processes during generation by using multi-token parallel generation and  n-gram retrieval for token reutilization, therefore enabling efficient ultra-long sequence generation~(up to 100K tokens).

\xhdr{Video-based MLLMs}
The necessity of token reduction primarily lies in enhancing the model's effective understanding of video content through:~\ding{172} \textit{Instruction-guided information filtering}: token reduction prioritizes selecting visual information relevant to user instructions over raw data volume.~\ding{173} \textit{preserving spatiotemporal structure}: token reduction strategically compresses massive spatiotemporal information to retain spatiotemporal dependencies, ensuring the model can capture dynamic semantics, as well as prevent redundant tokens interfere with long temporal reasoning.~\ding{174} \textit{Preserving semantic integrity}: it facilitates feasible processing of extremely long sequences in learning while preserving semantic integrity.~\ding{175} \textit{Multi-modal alignment}: token reduction distills visual information into a compact, semantically aligned form, thereby efficiently bridging the gap between language and vision~\cite{liu2025hybrid}, as shown in Fig.~\ref{fig:main_framework}f. By doing so, it effectively addresses the challenges posed by the low abstractness and lack of guidance inherent in raw visual inputs, which are the root causes of semantic misalignment and optimization ambiguity in multi-modal models.
Recent works illustrate these principles: HICom~\cite{liu2025hybrid} conducts conditional token compression at local and global levels using user instructions as guidance to retain instruction-relevant visual information while reducing computational burden. 
Video-XL-Pro~\cite{liu2025video} employs reconstructive token compression with a dynamic token synthesizer and semantic-guided masking to generate compact yet comprehensive video tokens for improved MLLM performance and efficiency.



\section{Future Directions}
\label{sec:future}

In this section, we outline eight future directions, from algorithmic
foundations (Sec.~\ref{sec_design}$\sim$~\ref {sec_hard}), through reasoning-centric methods
(Sec.~\ref{sec_in}$\sim$~\ref{sec_RL}), to real-world application domains (Sec.~\ref{sec_dense}$\sim$~\ref{sec_sci}),
tracing a path from core design principles to deployment at scale.

\subsection{Design of New Algorithms}
\label{sec_design}

Future research on algorithm design should explore holistic and adaptive token reduction strategies. Building on recent advances, we outline six promising directions:

\xhdr{Better Token Importance Metrics} 
It is critical to re-evaluate how token importance is defined and measured. More robust and unbiased scoring mechanisms can be developed, such as predictors~\cite{akhauri2025tokenbutler} or meta-learning frameworks that go beyond attention-based proxies. These models should capture downstream utility with minimal supervision, enabling adaptive pruning across tasks and domains.

\xhdr{Constructive Token Compression} Token reduction can shift from purely eliminative pruning to strategies that merge spatially or semantically similar tokens into compact summary vectors~\cite{li2024tokenpacker}.  

\xhdr{Mitigating Position Bias} In MLLMs, attention-based pruning methods~(e.g., FastV) often rely on attention scores from a fixed query token, leading to retained tokens concentrating in specific image regions~(e.g., lower corner)~\cite{wen2025token} with potential position bias. Future methods should preserve spatial diversity by enforcing structural uniformity in retained tokens to improve robustness on visual tasks.

\xhdr{Cross-Modal Guided Pruning} Pruning decisions in MLLMs should be guided by inter-modality dependencies, rather than made independently for each modality. For example, text-guided pruning of visual tokens can improve alignment between modalities~\cite{cao2024madtp}. The design should account for joint representations and semantic correspondence across all relevant inputs. 

\xhdr{End-to-End Sparsification} Token reduction should consider both the prefill stage and decoding phase for LLMs. This includes dynamically managing the sparsity of KV caches and selectively updating generated tokens, sustaining efficiency gains throughout the entire inference process ~\cite{huang2024dynamic}.

\xhdr{Hardware-Algorithm Co-Design} Token pruning can explore custom hardware and compiler optimizations that take advantage of dynamic token sparsity patterns~(e.g., irregular memory access and conditional computation) to maximize throughput and energy efficiency as detailed in Sec.~\ref{sec_hard}.

\subsection{Complementary to Other Methods}
\label{sec_com}
Beyond algorithmic design, token reduction can be further strengthened by composing it with orthogonal efficiency techniques, such as quantization. By selectively reducing the number of tokens processed during inference, models can improve both performance and efficiency, particularly when paired with quantization strategies~\cite{li2025mergevq}. Traditional key-value cache quantization methods often suffer from accuracy loss due to their inability to handle outlier tokens that carry distinct or rare features. To mitigate this issue, Outlier Token Tracking~\cite{su2025accuratekvcachequantization} identifies outlier tokens during decoding and excludes them from quantization, preserving full-precision necessary representations   and improving key-value cache quantization accuracy. 
Similarly, Agile-Quant~\cite{shen2024agile} incorporates token pruning as a preprocessing step to reduce the impact of activation outliers. It prunes tokens based on their attention to the start-of-sequence token, discarding those with low attentiveness, which often appear in adjacent channels and contribute to quantization noise. This targeted pruning reduces interaction distances between salient tokens and helps maintain model accuracy under low-bit quantization settings.

\subsection{Algorithm-Hardware Co-Design}
\label{sec_hard}
While algorithmic advancements in token reduction have achieved impressive computational savings, the next crucial step is to integrate these techniques with hardware-aware design principles.
We posit that algorithm-hardware co-design is essential for holistic optimization across the compute stack, considering the interplay between algorithmic choices, hardware architectures~(specialized data paths, memory hierarchies, communication fabrics, control logic, etc.), and compiler/runtime support~(efficient sparse mapping, dynamic scheduling, irregular-data management, etc.)~\cite{dong2023heatvit,parikh2024accelerating}.

Currently, co-design efforts targeted at token reduction lag significantly behind pure algorithmic research. This gap is problematic because hardware design needs to balance PPA~(power, performance, and area), platform specifics, data movement costs, control overhead, and scalability/reusability~\cite{10.1109/ASP-DAC58780.2024.10473968}. Algorithms developed in isolation often generate sparse or irregular compute patterns that general-purpose hardware cannot exploit effectively. 
Therefore, future research should aim to: 1) Design parameterizable, reconfigurable accelerator modules-such as on-the-fly importance-scoring units and sparse-data pipelines-that natively support token-reduced Transformers.  2) Explore Processing-in-Memory~(PIM) architectures to alleviate severe memory bottlenecks caused by dynamic token pruning. By executing scoring operations or partial attention mechanisms within or near memory arrays, PIM can drastically reduce data movement costs and improve end-to-end efficiency.

\subsection{From Prompt Tuning to Chain of Thought Reasoning}
\label{sec_in}
Current token reduction efforts for prompts have primarily aimed at compressing prompts for efficiency, often with impressive results~\cite{jiang2023llmlingua,mu2023learning}. Looking forward, token reduction should evolve into enhancing reasoning and maximizing utility per token in context. Instead of focusing solely on making prompts shorter, future research should explore how each remaining token can carry more information or trigger more complex inference during in-context learning and chain of thought reasoning.
One direction is to alter the generation paradigm itself, for example, training language models to predict multiple tokens per step~\cite{gloeckle2024better}. Another idea is to enable deeper internal reasoning without increasing prompt length~\cite{adams2023sparse}.

As mentioned in Sec.~\ref{sec:challenge}, long CoT chains can become verbose: excessive reasoning steps may introduce errors or obscure logical clarity, particularly in LLM agents where internal reasoning alternates with external tool use~\cite{gao2025txagent,wang2024toolgen}. Token reduction may serve a critical role in this context by compressing intermediate reasoning into a compact representation. For example, approaches based on next-token prediction~\cite{mu2023learning,zhang2025lightthinker} can distill intermediate thinking chains into a set of dense, information-rich tokens. These compressed representations can then replace the full intermediate context and serve as inputs for subsequent reasoning steps. 

This compressed-thinking strategy has two main benefits: 1) reducing error accumulation and keeping the logic clear by focusing on key information, and 2) allowing more reasoning rounds to fit within a fixed context window, enabling deeper multi-step inference without exceeding length limits.

In summary, the next phase of token reduction research should shift focus from simple prompt compression to reasoning-centric compression.  Rather than just trimming prompts, we should ask: \textit{How can we make each token in the prompt or context do more work for us?} This involves training models with objectives that reward higher-level inference per token, developing architectures that recycle tokens for multi-step thinking, or dynamically selecting the most salient tokens to keep at each step of reasoning. 

\subsection{Efficient Reasoning with Reinforcement Learning}
\label{sec_RL}
Reinforcement-learning (RL)-driven token reduction has shown strong promise for improving reasoning efficiency in both LLMs and MLLMs. The key challenge is to balance compute and reasoning quality via dynamic reward design, sparsity-inducing constraints, and adaptive control of effective token length~\cite{aggarwal2025l1,luo2025o1pruner,wu2025arm}.

In language reasoning, length awareness is incorporated either by diversifying reasoning formats during pre-training~\cite{su2025dualformer} or by adding explicit length penalties in the RL stage~\cite{arora2025,hou2025thinkprune}. While effective, most approaches implicitly assume static task complexity or rely on hand-specified length constraints, which can be suboptimal under heterogeneous workloads. A natural next step is \textit{adaptive reasoning}: using RL to learn per-instance budget allocation from intrinsic task difficulty~\cite{bercovich2025llamanemotron,qwq32b}. In parallel, performing reasoning in a compressed latent space can yield substantial computational savings~\cite{shen2025hiddenthinking}, but current methods often degrade due to poorly structured latent representations. A promising direction is to inject \textit{explicit logical structures} into the latent space to enable more controllable and compositional reasoning~\cite{helff2025}.

Beyond language modality, under the ``Fast-Slow Thinking'' framework~\cite{xiao2025fast}, RL can supervise hierarchical selection of high-value tokens: a fast branch applies sparsity signals (e.g., rule-based rewards or information-density scoring) to prune redundant visual/semantic features, while a slow branch allocates computation to refined reasoning. Additionally, RL enables a Think-with-Image paradigm~\cite{su2025thinking} by letting models adapt visual granularity: VisionThink~\cite{yang2025visionthink} adopts a progressive resolution strategy, using RL to selectively request high-resolution inputs only for fine-grained tasks like OCR. PixelThink~\cite{wang2025pixelthink} addresses visual overthinking in segmentation by adjusting reasoning length based on task difficulty. These methods demonstrate how RL can dynamically calibrate both input \textit{resolution} and reasoning \textit{depth} across various tasks. Looking ahead, integrating such approaches could enhance cross-modal alignment and inference efficiency in real-time and resource-constrained scenarios, supporting a new generation of lightweight yet capable multimodal language agents.

\subsection{Towards Dense Prediction Tasks for Vision}
\label{sec_dense}
Existing works primarily concentrate on compressing the backbone of models to ensure their generalization ability, and few works explore recovering all tokens for dense prediction tasks~\cite{bolya2023token,liang2022expediting}. 
It is necessary to develop custom token reduction methods for various downstream dense prediction applications like autonomous driving and robotic control with specific settings and requirements~\cite{cao2025fastdrivevla,pei2025action}, as shown in Fig.~\ref{fig:main_framework}g.
Lacking these specialized designs would lead to a mismatch and performance drop when deployed in real-world settings. For example, autonomous driving~\cite{zhang2024sparsead,peng2025colavla,xiong2025prune2drive,cao2026fastdrivevla} would require displacement and velocity based on occupancy prediction, and robotic control~\cite{wang2024sparse,yang2025efficientvla,feng2026gridsampler} would demand rotation angle according to the grid map. Therefore, developing fast and specialized token reduction strategies tailored for downstream dense prediction tasks is crucial for deployment in practical scenarios.

\subsection{Towards Long Video Applications}
\label{sec_long}
Exploiting long videos holds great potential, as processing hours of footage is significantly more labor-intensive and time-consuming than working with short clips. Due to the inherent complexity and resource demands, most current research on long video learning focuses on discriminative tasks such as video understanding~\cite{lee2024video, ren2023testa,chen2026streamingtom}.
In contrast, broader applications including long video editing~\cite{zhang2025adaflow}, long video-text retrieval, and narrative-level generation remain largely underexplored. 
Progress in these areas could have a significant impact on scene editing in video clips, character rendering in movies, and retrieving useful information from numerous videos.

Moreover, token reduction offers a path toward interpretability and efficiency in long video processing~\cite{jiang2025token}. This mimics the human visual system, which does not attend to every frame in detail but instead focuses on salient spatiotemporal changes, such as actions or object movement, while filtering out static, redundant content like backgrounds and stationary objects. Future models should similarly prioritize informative frames and temporal segments, allowing them to reason over extended video sequences with greater efficiency and interpretability.

\subsection{Towards Efficient Agentic Systems}

\xhdr{Dynamic Memory \& Context Engineering} 
AI agents face a severe context bottleneck where accumulating interaction history not only incurs quadratic computational costs but also degrades reasoning through "lost-in-the-middle" phenomena. Transitioning to active context management is essential~\cite{anthropic2025effective}; systems must distinguish between immutable instructions and transient episodic data. ACON~\cite{kang2025acon} demonstrate that semantic memory compression can reduce peak token usage while maintaining long-horizon performance. AgentOCR~\cite{feng2026agentocr} shows that semantic memory compression can reduce token usage while maintaining long-horizon performance. Complementary strategies include hierarchical memory abstraction, which offloads older history to retrieval-based storage, and hybrid masking techniques~\cite{jetbrains2025efficient}. These approaches treat tokens as a finite resource to be optimized on-the-fly, dynamically adjusting the sparsity of the context window based on the complexity of the current reasoning step.

\xhdr{Observation \& Tool Pruning} 
Agents must often process verbose tool outputs (e.g., massive JSON/HTML). Feeding raw data is inefficient; future research should focus on token-aware interaction, where lightweight scorers filter observations to retain only task-relevant features. This preserves critical context bandwidth without sacrificing trajectory accuracy, as evidenced in recent programming agents~\cite{xiao2025improving}. Furthermore, adaptive truncation strategies that prioritize error traces over standard success logs can significantly improve debugging capabilities while minimizing token consumption, as shown in Fig.~\ref{fig:main_framework}h.

\xhdr{Communication-Efficient Multi-Agent Systems} 
At the system level, the aggregate token cost scales with the number of interacting nodes. Enforcing strict token budgets on inter-agent exchange compels information-dense communication. This sparse protocol enhances scalability and accuracy by reducing redundant chatter and mitigating hallucinations in collaborative workflows~\cite{zhang2024cut,zou2025latent,bai2026aiagentsspendmoney}.

\subsection{Towards AI for Broader ML and Scientific Domains}
\label{sec_sci}
Token reduction methods can also offer powerful opportunities to reshape broader machine learning and scientific applications. In particular, domains such as medicine, biology, chemistry, and temporal data analysis frequently encounter complex data structures, heterogeneous data sources, and intricate domain-specific relationships. Informed tokenization approaches promise to address these challenges by transforming complex and rich scientific data into concise, informative, and flexible representations, significantly enhancing the utility of transformer-based foundation models across these domains.

\xhdr{Building Biomedical Tokenizers}
Recent works exemplify the transformative potential of advanced tokenization methods in the biomedical domain, including protein~\cite{gao2025foldtoken,suyunu2025evobpe,yuan2025protein}, genomic~\cite{eapen2025genomic}, and chemical structure~\cite{yan2024invariant} tokenizers. Collectively, these methods illustrate how informed reduction and condensation of input tokens can lead to more effective and interpretable scientific models. 
For example, traditional tokenizers in EHR foundation models typically treat medical codes as isolated textual units, neglecting their inherent structured and relational context, such as hierarchical relationships, disease co-occurrences, and drug-treatment associations found within biomedical ontologies.
To solve this issue, MedTok~\cite{su2025multimodal} integrates textual descriptions and graph-based relational data into a unified tokenization framework. 
It first uses a language model encoder to extract embeddings from medical code descriptions and employs a graph encoder to capture relational structures from biomedical ontologies. These embeddings are combined into a compact token space through vector quantization, preserving both modality-specific and cross-modality information. 

To enhance informativeness and reduce redundancy, MedTok employs a token packing mechanism. It optimizes shared tokens and modality-specific tokens, ensuring that the final tokens encode both shared semantic meaning and modality-specific structure. This process drastically reduces effective vocabulary size, addressing the scalability challenge of 600,000+ medical codes by collapsing redundant representations while preserving critical clinical context.
Inspired by adaptive tokenization methods for vision~\cite{duggal2024adaptive,kang2023tictok,yan2024elastictok}, future EHR tokenization would be adaptive, enabling the dynamic representation of patients' medical histories, where the length of the token series for each patient's history would be directly correlated with the length and complexity.
Such adaptive tokenization can significantly improve training and inference efficiency across diverse healthcare systems. 

\xhdr{Time-Series Data and Clinical Reasoning}
Temporal dynamics form an essential component of clinical reasoning, particularly through longitudinal patient data like lab tests and vital signs. However, current large language models struggle to effectively incorporate time-series inputs due to challenges in temporal tokenization~\cite{spathis2024first,anjum2024lipcot,fang2025tsla,spathis2024first,masserano2024enhancing}.
Future tokenization methods should not only dynamically adjust the number of tokens according to temporal complexity but also selectively focus on time segments most relevant to the clinical context, prompt, or task at hand~\cite{talukder2024totem}.
This could enhance training effectiveness and inference accuracy, helping create the next generation of EHR foundation models, which are flexible not only over different tasks or prompts, but also over different data sources, patients, and populations. 
The complexity and richness of EHR data offer opportunities for AI-driven advancements in patient health outcomes. Future EHR models should support comprehensive reasoning capabilities, encompassing complete patient histories, such as vitals, lab results, diagnoses, and procedures over time. They could facilitate timely disease predictions, accurately forecast chronic disease trajectories, and anticipate patient responses to treatments. 

\section{Conclusion}
\label{sec:conclusion}
In this position paper, we have argued that token reduction must evolve beyond a mere efficiency optimization to become a core design principle in generative modeling. We have shown how principled token reduction can address key challenges such as enhancing semantic fidelity in vision-language alignment, curbing verbose reasoning trajectories, preserving long-range coherence, and stabilizing learning dynamics. 
Looking forward, the roadmap we outlined points to a broad landscape of opportunities, ranging from algorithmic innovations and hardware-algorithm co-design to specialized applications in scientific domains. We anticipate that future work will increasingly focus on constructive compression and reinforcement learning-guided selection, enabling models to autonomously optimize their information bandwidth. Ultimately, by treating token reduction as a holistic and task-aware mechanism, the community can develop next-generation systems that effectively balance scalability with effectiveness, interpretability, and performance.

\section{Limitations}
\label{sec:limitations}
While token reduction offers significant benefits, our review identifies critical limitations and trade-offs that must be considered to avoid indiscriminate application.

\paragraph{Information Loss in Dense Prediction}
Token reduction methods, particularly pruning, inherently discard information. While this is acceptable for semantic classification or generation, it poses severe risks for dense prediction tasks (e.g., segmentation, object detection) or medical analysis where fine-grained spatial details are crucial. 
Merging strategies like ToMe~\cite{bolya2023token} mitigate this better than pruning, but artifacts often remain at high compression ratios. 
In scenarios requiring pixel-perfect reconstruction, the trade-off between reduction and precision often favors preserving the full token set. Furthermore, the lack of specialized designs for token recovery often leads to performance mismatches in real-world applications like autonomous driving and robotic control. Future research must therefore explore custom reconstruction mechanisms to ensure these methods can meet the rigorous demands of dense tasks.

\paragraph{Overhead vs. Gain}
Dynamic token reduction introduces computational overhead (e.g., scoring networks, predictors). 
For short sequences or small batch sizes, the cost of computing token importance may outweigh the savings from processing fewer tokens. Furthermore, unstructured pruning can lead to irregular memory access patterns that are inefficient on standard hardware (GPUs/TPUs), potentially negating theoretical FLOPs reductions.

\paragraph{Alternatives: Reduction vs. Retrieval}
Critics may argue that techniques like Retrieval-Augmented Generation (RAG) or simply scaling context windows render token reduction unnecessary. 
However, we argue these are complementary. 
While RAG selects \textit{documents}, token reduction operates at the \textit{sub-document} level, filtering noise within relevant chunks. 
Similarly, while larger context windows allow for more input, they exacerbate the "lost-in-the-middle" phenomenon; token reduction acts as an attention-sharpening mechanism that helps models utilize these long contexts more effectively.


{
    \small
    \bibliographystyle{plain}
    \bibliography{reference}
}


\newpage
\appendix

\clearpage
\newpage
\section{Theoretical Formulation}
\label{sec:appendix_methodology}

In this section, we provide a unified mathematical framework for token reduction methods. We formalize the token reduction process into two distinct phases: \textit{Compression Criteria} (how to evaluate tokens) and \textit{Compression Strategies} (how to reduce tokens).

\subsection{Problem Definition}
Given an input sequence $X = [x_1, x_2, \dots, x_N] \in \mathbb{R}^{N \times D}$, where $N$ represents the sequence length and $D$ denotes the feature dimension, the objective of token reduction is to generate a compressed sequence $X' = [x'_1, x'_2, \dots, x'_M] \in \mathbb{R}^{M \times D}$, such that $M \ll N$.

The reduction process can be formally defined as a composite function of a scoring criterion $\mathcal{E}$ and a compression strategy $\mathcal{P}$:
\begin{equation}
    X' = \mathcal{P}(X, \mathcal{E}(X))
\end{equation}
where $\mathcal{E}(X)$ outputs importance scores, clustering assignments, or gradient sensitivities, and $\mathcal{P}$ executes the dimensionality reduction. Fig.~\ref{fig:token_method} shows the token reduction pipeline.

\subsection{Compression Criteria}
The scoring function $\mathcal{E}: X \to \mathcal{S}$ determines the semantic value or redundancy of each token. We broaden the categorization to include gradient and entropy-based metrics alongside standard parametric/non-parametric approaches.

\xhdr{Attention-based Scoring}
Utilizing the inherent sparsity of the self-attention mechanism, the importance of a token $x_i$ is quantified by the attention it receives. This can be \textit{global} (averaged across all heads/tokens) or \textit{targeted} (attention from the special [CLS] token or specific query tokens). The score $s_i$ is typically calculated as:
\begin{equation}
    s_i = \sum_{j \in \mathcal{Q}} \text{Attn}(x_j, x_i)
\end{equation}
where $\mathcal{Q}$ is the set of query tokens (e.g., $\mathcal{Q}=\{x_{\text{CLS}}\}$ for classification tasks or $\mathcal{Q}=\{x_{1 \dots N}\}$ for global density).

\xhdr{Similarity-based Scoring}
This approach assumes that tokens close in the feature space contain redundant information. The criterion calculates pairwise distances to identify clusters or redundant pairs. For tokens $(x_i, x_j)$, the metric is typically cosine similarity:
\begin{equation}
    \text{Sim}(x_i, x_j) = \frac{x_i \cdot x_j}{\|x_i\|_2 \|x_j\|_2}
\end{equation}
High similarity scores ($\text{Sim} > \tau$) trigger merging operations. Advanced methods extend this to density-based clustering (e.g., K-Means or DPC-KNN) to identify representative centroids.

\xhdr{Gradient and Entropy-based Scoring}
Beyond static feature analysis, recent methods employ dynamic metrics. \textit{Gradient-based} criteria measure a token's contribution to the loss function, retaining tokens with high gradient norms ($\| \nabla_{x_i} \mathcal{L} \|$). \textit{Entropy-based} criteria evaluate the uncertainty of the model's prediction; tokens with low information density (low entropy) are candidates for pruning in early-exit or fast-forwarding frameworks.

\xhdr{Parametric Scoring}
Parametric methods introduce a lightweight auxiliary module (e.g., a predictor network $\mathcal{M}_{\phi}$) to explicitly predict token utility:
\begin{equation}
    S = \mathcal{M}_{\phi}(X)
\end{equation}
where $S \in [0,1]^N$ represents the keep-probability or saliency score. These predictors are trained via Gumbel-Softmax or reinforcement learning (RL) to maximize downstream accuracy while minimizing token count.

\subsection{Compression Strategies}

\begin{figure*}[t]
\vspace{-0.4cm}
  \centering
  \includegraphics[width=1.0\linewidth]{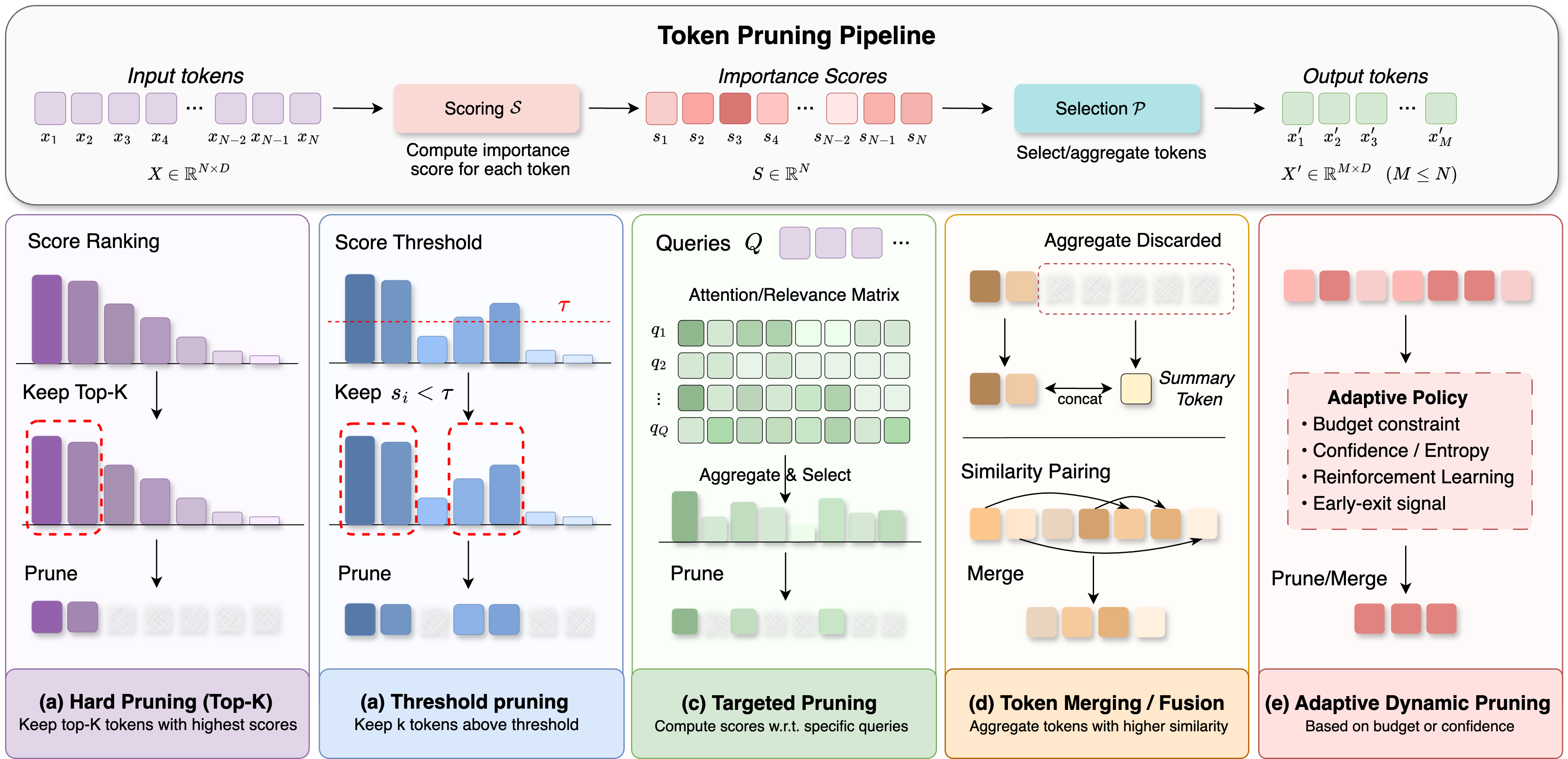}
  \vspace{-0.4cm}
  \caption{\textbf{Visualization of difference token reduction techniques.}  
Given an input token sequence, token pruning first estimates token utility using a scoring criterion, such as attention relevance, task-specific query relevance, entropy, confidence, or a learned predictor. 
Based on these scores, different selection strategies can be applied, including top-$K$ pruning, threshold-based pruning, targeted query-based pruning, soft pruning with summary-token aggregation, and adaptive pruning with dynamic token budgets. }
  \label{fig:token_method}
\end{figure*}

Once the relationships or scores are established, the compression strategy $\mathcal{P}$ transforms the sequence. We classify these into four primary mechanisms.

\xhdr{Token Pruning (Hard \& Soft)}
Pruning is a selection process that discards tokens based on the criteria $\mathcal{E}$.
\begin{equation}
    X' = \{ x_i \mid s_i \in \text{TopK}(S) \lor s_i > \tau \}
\end{equation}
While \textit{Hard Pruning} permanently removes tokens, \textit{Soft Pruning} (or "packaging") aggregates the discarded tokens into a single summary token to preserve residual information, preventing total information loss.

\xhdr{Token Merging \& Clustering}
Merging aggregates information from a set of tokens $\mathcal{C} = \{x_1, \dots, x_k\}$ identified as similar. This ranges from bipartite matching (merging pairs) to density-based clustering (merging large groups). The merged token $x'_{\text{cluster}}$ is a weighted average:
\begin{equation}
    x'_{\text{cluster}} = \frac{\sum_{x_j \in \mathcal{C}} w_j x_j}{\sum_{x_j \in \mathcal{C}} w_j}
\end{equation}
where $w_j$ tracks the token's size (number of constituent patches), ensuring proportional representation of fused features.

\xhdr{Transformation-based Compression}
These methods reduce sequence length through structural operations, exploiting spatial (image) or temporal (video) priors. Common techniques include:
\begin{itemize}
    \item \textbf{Pooling/Unshuffle:} Non-parametric downsampling: $\text{Pool}: \mathbb{R}^{H \times W} \to \mathbb{R}^{\frac{H}{r} \times \frac{W}{r}}$.
    \item \textbf{Convolution:} Strided convolution to abstract local neighborhoods: $X' = \text{Conv}_{k \times k}(X, s)$.
\end{itemize}

\xhdr{Token Distillation (Query-based)}
Distillation employs a set of learnable latent queries $Q \in \mathbb{R}^{M \times D}$ to extract information from the input $X$ via cross-attention mechanisms (e.g., Perceiver Resampler or Q-Former). This decouples output length $M$ from input length $N$:
\begin{equation}
    X' = \text{Softmax}\left(\frac{Q (X W_K)^T}{\sqrt{D}}\right) (X W_V)
\end{equation}
This strategy is particularly effective for cross-modal alignment, compressing dense visual features into sparse text-aligned tokens.

We summarize the general workflow of training-free token reduction in Algorithm \ref{alg:token_reduction}. We also show a visualization of token reduction in Fig.~\ref{fig:token_method}.

\begin{algorithm}[h]
\caption{General Training-Free Token Reduction Workflow}
\label{alg:token_reduction}
\begin{algorithmic}[1]
\Require Input tokens $X \in \mathbb{R}^{N \times D}$, Target ratio $r$
\Ensure Compressed tokens $X' \in \mathbb{R}^{M \times D}$
\State $M \gets \lfloor N \times (1-r) \rfloor$
\State \textbf{Phase 1: Criteria Calculation ($\mathcal{E}$)}
\If{\textit{Attention-based}}
    \State $S \gets \text{Agg}(\text{AttentionMap}(X))$ \Comment{Agg: Sum/Avg over heads}
\ElsIf{\textit{Similarity-based}}
    \State $A \gets X X^T / (\|X\| \|X\|)$ \Comment{Cosine Similarity Matrix}
    \State Partition $X$ into sets $\{\mathcal{C}_1, \dots, \mathcal{C}_M\}$ via clustering on $A$
\EndIf
\State \textbf{Phase 2: Strategy Execution ($\mathcal{P}$)}
\If{\textit{Pruning}}
    \State $\text{Indices} \gets \text{TopK}(S, M)$
    \State $X' \gets \text{Gather}(X, \text{Indices})$
\ElsIf{\textit{Merging}}
    \For{$m = 1$ to $M$}
        \State $x'_m \gets \text{WeightedSum}(\mathcal{C}_m)$
    \EndFor
\EndIf
\State \Return $X'$
\end{algorithmic}
\end{algorithm}

\subsection{Controllable Reasoning via Reinforcement Learning}
Controllable Reasoning refers to the ability of language models to dynamically adjust the depth and length of their reasoning processes according to user-specified constraints (e.g., exact or maximum token lengths), thereby enabling a tunable trade-off between reasoning efficiency and accuracy. State-of-the-art approaches~\cite{luo2025o1pruner,aggarwal2025l1} typically employ reinforcement learning frameworks~\cite{shao2024deepseekmath}, where a multi-objective reward function is designed to optimize correctness and length compliance jointly. Specifically, the reward function incorporates two key objectives:
\begin{enumerate}
    \item \textbf{Correctness reward}: awarded if the model's output matches the ground-truth answer;
    \item \textbf{Length penalty}: imposed if the generated sequence length deviates from the target length.
\end{enumerate}

\noindent \textbf{Exact Length Control.} It requires the model to generate reasoning sequences whose length exactly matches a user-specified target length:
\begin{equation}
R_{\text{exact}}(y, y_{\text{t}}, n_{\text{t}}) = \mathbb{I}(y = y_{\text{t}}) - \alpha \cdot |n_{\text{t}} - n_y|,
\end{equation}

where $y$ is the generated sequence, $y_{\text{t}}$ is the ground truth answer, $n_{\text{t}}$ is the target token length, $n_y$ is the actual token length of the generated sequence, $\mathbb{I}(\cdot)$ is the indicator function (1 if correct, 0 otherwise), and $\alpha$ is a penalty weight balancing correctness and length.

\noindent \textbf{Maximum Length Control.} It controls the model to generate reasoning sequences no longer than a specified upper limit, encouraging efficient reasoning within a token budget:
\begin{equation}
R_{\text{max}}(y, y_{\text{t}}, n_{\text{t}}) = \mathbb{I}(y = y_{\text{t}})\cdot\text{clip}\left( \alpha \cdot (n_{\text{t}} - n_y) + \delta, 0, 1 \right),
\end{equation}

where $\text{clip}(\cdot, 0, 1)$ clamps reward to the range $[0, 1]$ and $\delta$ is an offset term to avoid zero reward (typically set to 0.5).

\noindent \textbf{Length Efficiency Optimization.} It encourages the model to shorten reasoning length while maintaining correctness, particularly useful for reducing redundancy in long-reasoning models:
\begin{equation}
R_{\text{eff}}(y, y_{\text{t}}, x) = \frac{\bar{L}_{\text{ref}}(x)}{L(y)} - 1 + \lambda \left( A(y, y_{\text{t}}) - \bar{A}_{\text{ref}}(x) \right),
\end{equation}

where $\bar{L}_{\text{ref}}(x)$ is the average length of reference model outputs for problem $x$, $A(y, y_{\text{t}})$ is the accuracy function (1 if correct, 0 otherwise), $\bar{A}_{\text{ref}}(x)$ is the average accuracy of the reference model on problem $x$, and $\lambda$ is an accuracy penalty weight to prevent performance degradation from over-compression.

By adjusting hyperparameters (e.g., $\alpha$ and $\lambda$) in the reward function, one can flexibly control the model's tendency to prioritize correctness versus length, which not only achieves precise length control but also maintains or even improves model performance while significantly reducing reasoning overhead.


\end{document}